%% file: main.tex
\documentclass[runningheads]{llncs}
\pdfoutput=1 

\usepackage[utf8]{inputenc}
\usepackage{hyperref}
\usepackage[misc]{ifsym}
\usepackage{graphicx}
\usepackage{wrapfig}
\usepackage{verbatim}
\usepackage{paralist,amsmath,color,enumerate,framed}
\usepackage{todonotes}
\usepackage{multicol}

\title{Commonsense Ontology Micropatterns}
\author{Andrew Eells,\inst{1} Brandon Dave,\inst{2} Pascal Hitzler,\inst{1} Cogan Shimizu\inst{2}}
\institute{Kansas State University, Manhattan, KS, USA \and Wright State University, Dayton, OH, USA}
\authorrunning{Eells, Dave, Hitzler, Shimizu}
\begin{document}
\maketitle
\input{abstract}
\section{Introduction}
\label{sec:intro}
Humans perform transfer learning through analogous connections in understanding the unknowns of the world with previously acquired knowledge \cite{ar-1,ar-2}. In the scope of problem-solving, humans find similarities in their current problem to draw logical connections that may aid in finding a working solution, which, regardless of success, becomes part of the learning process for the next iteration of problem-solving. The task of drawing similarities requires humans to utilize their logical skills for generalizing a problem or concept to allow for adaptations from their acquired knowledge. 

Neurosymbolic Artificial Intelligence (AI) systems, in their pursuit of General AI, also seek this ability to conduct transfer and analogy learning.
We posit that the use of modular ontology is one way that this can be accomplished \cite{modont}. In particular, the notion of starting with a simple and (perhaps) well-understood pattern of knowledge and adapting it to new scenarios and use-cases; the mechanism by which we accomplish this we call \emph{modularization}. This overarching methodology is known as Modular Ontology Modeling (MOMo), and it is by now an established paradigm for constructing knowledge graphs (KG) and their schemas (i.e., an ontology) \cite{momo-swj}.

Ontology design patterns, which are tiny ontologies that are intended to solve domain-invariant modeling problems \cite{odp1}, are the usual starting point for the MOMo methodology. They may be sourced from many places, including the Ontology Design Pattern Portal\footnote{\url{https://ontologydesignpatterns.org/}} or Modular Ontology Design Libraries (MODLs) \cite{modl}. MODLs are annotated, curated collections of ODPs, combined with an index ontology that can be queried for patterns of different types and relationships between patterns.

In this paper, we provide such a design library, which we call CS-MODL, composed of \emph{commonsense micropatterns}. For the purposes of this work, we define our title.
\begin{compactitem}
    \item \textbf{commonsense}\quad we loosely mean ``what first comes to mind.'' In this case, we ask an LLM about common properties of common nouns (as discussed below) and receive some simple and naïve responses, which are then encoded into: 
    \item \textbf{micropatterns}\quad we contrast micropatterns versus ODPs. In this case, micropatterns do not have a sophisticated or rich semantics. Indeed, they are defined using \textsf{rdfs:domain} and \textsf{rdfs:range}, without more complicated OWL axioms. 
\end{compactitem}
Concretely, we describe the methodology we used to construct the 104 commonsense micropatterns contained in CS-MODL, discuss an example, and provide the library as an online resource.

The rest of the paper is organized as follows:
Section \ref{sec:meth} introduces our methodologies for obtaining the appropriate information from our chosen LLM and constructing the commonsense micropatterns. 
Section \ref{sec:example} provides an example of a generated commonsense micropattern.
Section \ref{sec:csmodl} summarizes the contributions of our commonsense library of patterns.
Section \ref{sec:conc} wraps up the paper with notes on future areas we observe as next steps to this research.

\section{Methodology}
\label{sec:meth}
The construction of these commonsense patterns was completed in two steps. First, we prompt an LLM in several ways (Section \ref{ssec:Prompts}) to produce an ontology design pattern for each noun in question (Section~\ref{ssec:nouns}). Then, the responses are consolidated into patterns via a set of heuristics (Section~\ref{ssec:resp}), and the patterns are packaged into a MODL (Section~\ref{sec:csmodl}).

\subsection{Large Language Models as Commonsense}
\label{ssec:llm}
LLMs, like GPT-4, ingest and learn from more information than any human possibly could. While it is certainly arguable that any LLM demonstrates human-level intelligence, it is also inarguable that it can present a commonsense view of the world -- in particular through the a declarative memory, having learned, perhaps as an accident of learning from ``mostly correct'' language, facts about the world \cite{llm-crawling}.

They are able to represent this information and ``knowledge'' in human\--readable ways with surprising fluency, and as we demonstrate below: frequently in valid RDF \cite{rdf-tr}. The following sections describe how we decided on which facts to extract (i.e., which commonsense notions to prompt), how we prompted for them, and the resultant collection of patterns.

\subsection{Noun Selection}
\label{ssec:nouns}
The nouns selected are the 101 most common nouns in American English, as listed in the Corpus of Contemporary American English (COCA) \cite{EnglishCorporaCOCA}. We additionally include the nouns ``chair", ``sofa", and ``loveseat" from our preliminary explorations for viability of our idea. All 104 nouns in our investigation are listed in Figure~\ref{fig:nouns}. 

\begin{figure}[t]
\begin{framed}
\begin{multicols}{4}
\begin{compactenum}
    \item Air
    \item Area
    \item Art
    \item Back
    \item Body
    \item Book
    \item Business
    \item Car
    \item Case
    \item Chair
    \item Change
    \item Child
    \item City
    \item Community
    \item Company
    \item Couch
    \item Country
    \item Day
    \item Door
    \item Education
    \item End
    \item Eye
    \item Face
    \item Fact
    \item Family
    \item Father
    \item Force
    \item Friend
    \item Game
    \item Girl
    \item Government
    \item Group
    \item Guy
    \item Hand
    \item Head
    \item Health
    \item History
    \item Home
    \item Hour
    \item House
    \item Idea
    \item Information
    \item Issue
    \item Job
    \item Kid
    \item Kind
    \item Law
    \item Level
    \item Life
    \item Line
    \item Lot
    \item Loveseat
    \item Man
    \item Member
    \item Minute
    \item Moment
    \item Money
    \item Month
    \item Morning
    \item Mother
    \item Name
    \item Night
    \item Number
    \item Office
    \item Others
    \item Parent
    \item Part
    \item Party
    \item People
    \item Person
    \item Place
    \item Point
    \item Power
    \item President
    \item Problem
    \item Program
    \item Question
    \item Reason
    \item Research
    \item Result
    \item Right
    \item Room
    \item School
    \item Service
    \item Side
    \item Sofa
    \item State
    \item Story
    \item Student
    \item Study
    \item System
    \item Teacher
    \item Team
    \item Thing
    \item Time
    \item War
    \item Water
    \item Way
    \item Week
    \item Woman
    \item Word
    \item Work
    \item World
    \item Year
\end{compactenum}
\end{multicols}
\end{framed}
\caption{The 104 nouns conceptualized into our commonsense micropatterns.}
\label{fig:nouns}
\end{figure}

\subsection{Prompt Engineering}
\label{ssec:Prompts}
We prompted the LLM\footnote{For this work, we used GPT-4-0613.} in several different ways, as we wished to test how it would respond to multiple variants of the same prompts. The initial prompts were a command (\ref{1}) and a request (\ref{2}), and then variations were added to include the noun as ``the following" (\ref{3} and \ref{4}), as seen below.

\begin{compactenum}\label{InitialPrompts}
    \item \label{1} Generate an ontology that covers [noun].
    \item \label{2} Could you develop a basic design pattern for representing [noun] in an ontology?
    \item \label{3} Generate an ontology that covers the following: [noun].
    \item \label{4} Could you develop a basic design pattern for representing the following in an ontology: [noun].
\end{compactenum}
Finally, the LLM was prompted to generate 9 additional variations of each prompt. See below for Prompt \ref{1}'s variations. In each case, the prompt was tried with and without the additional instruction of ``Provide it in valid Turtle/RDF format, excluding any extra text.'' for a total of 20 prompts per noun.

\begin{enumerate} \label{PromptsGroup1}
    \item Generate an ontology that covers [noun].
    \item Develop an ontology dedicated to [noun].
    \item Construct an ontology focused on [noun].
    \item Build an ontology surrounding the concept of a [noun].
    \item Formulate an ontology related to [noun].
    \item Establish an ontology based on [noun].
    \item Design an ontology to encompass [noun].
    \item Produce an ontology specifically for [noun].
    \item Compose an ontology to represent [noun].
    \item Make an ontology that pertains to [noun].
\end{enumerate}

\subsection{Response Consolidation}
\label{ssec:resp}
As noted, the LLM generated 80 responses for each of the 104 nouns in our collection. These responses were stored in a tab-delimited (TSV) file. However, not every response contained valid RDF and some contained no RDF at all. The number of responses was too large for manual processing, so we developed a script that would apply heuristics to each response file in order to integrate the responses into a single micropattern. The script consists of three parts, applied to each noun:
\begin{compactenum}
    \item \textbf{extraction} -- for each specific prompt response (i.e., a row in the TSV), the script attempts to identify and extract the portion of the response in RDF. Unfortunately, not every response output an ODP or easily recognizable RDF. Generally speaking, cleaning required the removal of extra text (even when prompted \emph{not} to provide extra text) and surrounding Markdown formatting (e.g., code fencing).
    
    \item \textbf{cleaning} -- a set of heuristics was applied to each extract RDF (generally in turtle format) to allow for ingestion of the RDF into an \textsf{rdflib}\footnote{\url{https://rdflib.readthedocs.io/en/stable/}} graph structure. The most common type of cleaning necessary was the usage of newline characters. The LLM understood how to comment (i.e., separating class declarations from properties from assertions) but did not understand new lines. This frequently resulted in malformed Turtle, since the first part of the triple would be commented out.
    
    \item \textbf{integration} -- once each valid (or cleanable) TTL file was extracted from the response, we combine the properties (removing all triples pertaining to individuals), and annotate the result, giving us a commonsense micropattern for the noun. The annotations are further discussed in Section~\ref{sec:csmodl}.
\end{compactenum}
While we note that we combine all (valid and cleanable) responses into a single micropattern, the script is set up in a much more flexible way.

First, it is necessary to recall that the process by which ODPs are utilized in the MOMo process is based on customization to the use-case at hand; that is properties are frequently removed, added, or renamed for best fit. This process is known as \emph{template-based instantiation} \cite{template}, and it is a core aspect of the methodology. Specifically, each property with (explicit or inferred) domain and range, is assessed for fit. If it is not applicable (i.e., out of scope or too specific or general), the property and associated sub-graphs are removed. Conversely, if there are parts of the use-case that are not covered by the ODP, these must be added. In automated processes, it is generally easier to pare down an overly superfluous pattern. That is, if there are no matches or fillers for the slots that a pattern provides, they should be dropped. We follow this strategy for our micropatterns, meaning that we include each property from every response, which includes variants of the same property. For example, in Figure~\ref{fig:example}, which shows a schema diagram of a commonsense micropattern for \texttt{Air}, \textsf{hasHumidity} has a range of both \textsf{Humidity} and \textsf{xsd:float}. It is conceivable that one or the other is unnecessary, or that they may be used in tandem (i.e., where a connection from \textsf{Humidity} is made to the \textsf{xsd:string}), e.g., as a type of shortcut or view \cite{owled-typecasting,FOIS-KWG-Lite}.

Yet, this strategy may not be applicable for all systems. As such, the script allows for a voting mechanism to take place. This might be useful if additional prompt variants are used or if multiple LLMs are prompted for ODPs, thus allowing for properties (with domains and ranges) that appear in multiple patterns from multiple sources are those that appear in the final micropattern. The voting mechanism is as follows.

For the usable response, we extracted 3-tuples expressing Properties associated with a noun, and the Property's respective described scoped domain and range. This process also included the notion of a voting process, which would decide the final noun pattern's implemented properties based on occurrence per generated noun ontology. Our noun ontologies ended up with limited duplicate occurrences of the same triplets (s, p, o) (ranging from 1 to 4 total occurrences), and so, we chose not to implement the voting process and a threshold to implement properties to the final noun pattern. This allows us to maximally express the noun's ontology as stripping the ontology down would be an easier task than building around a concept. 

Finally, when generating the micropattern, we minimally describe the properties and classes. That is, we use only \textsf{rdfs:Class}, \textsf{rdfs:subClassOf}, \textsf{rdfs:range}, and \textsf{rdfs:domain}. The script, however, is designed in such a way to be easily modifiable to improve the semantic expressivity of the patterns. For example, we could use a simple heuristics, plus a set of axiom patterns (i.e., taken from \cite{owlaxax}), to assert the most common types of constraints for a particular property.

\section{Example Pattern}
\label{sec:example}
\begin{figure}[t]
    \centering
    \includegraphics[scale=0.5]{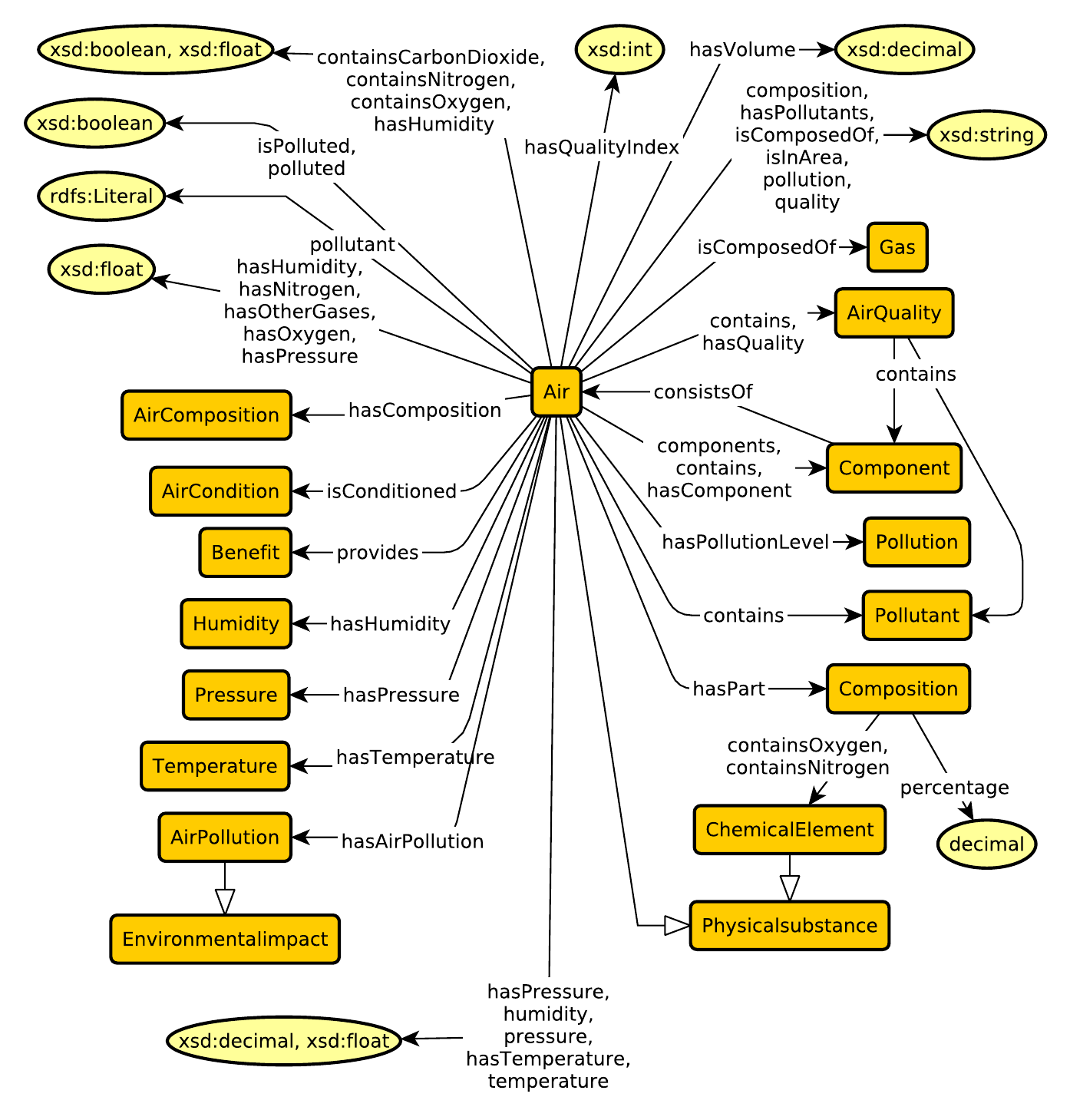}
    \caption{A graphical schema diagram of the Air ontology}
    \label{fig:example}
\end{figure}
We have selected a pattern, \texttt{Air}, which exemplifies many interesting properties of our commonsense micropatterns. The schema diagram for this micropattern is shown in Figure~\ref{fig:example}. The gold boxes reflect classes. The yellow ellipses indicate datatypes. Black filled arrows are binary relations. When labels overlap multiple arrows, we use this to show that there are multiple ranges for the property.

As noted in Section~\ref{ssec:resp}, there are some reuse of property names (e.g., \textsf{hasHumidity}) that would technically be not allowed in OWL. However, we suggest that \emph{pruning} is easier than growing, so a system that utilizes CS-MODL can simply remove the unused property, as necessary.

The turtle file for this example can be found in the repository.

\subsection{License \& Availability}
\label{ssec:avail}
Our artifacts are made available online through a GitHub repository.\footnote{\url{https://github.com/kastle-lab/commonsense-micropatterns/}}
All data (which includes documentation, prompts, responses, extracted and cleaned TTL files, integrated patterns, and CS-MODL itself) and the supporting scripts that generated the data are provided under an open and permissive license (GNU Lesser General Public License v2.1), and it is also included explicitly in the repository.

\section{A Commonsense Modular Ontology Design Library}
\label{sec:csmodl}
The purpose of a MODL is to collect together (curate) a set of ontology design patterns. In the first version \cite{modl}, these were generally well-documented and useful patterns. A forthcoming second version has many more, spanning a wider set of application scenarios.\footnote{\url{https://github.com/kastle-lab/modular-ontology-design-library}} This particular MODL, CS-MODL, is similar insofar that it also satisfies broad use-cases. The purpose of which is to collect together patterns of similar quality, abstraction, and expressivity for the purpose of modeling commonly occurring entities (e.g., in informal language or the real world).

The CS-MODL, like other MODLs, is programmatically queryable through OPaL annotations \cite{opla,opla-cp} indicating that CS-MODL contains sufficient metadata scaffolding such that top-level conceptual components (i.e., the key notion) of patterns can be easily matched to external entities detected by a framework.\footnote{We note that we are trying to have a better acronym -- OPaL -- instead of the previously published acronym!} For example, if a vision interpreter finds something with four legs, it could match to the \texttt{Chair} pattern. Inversely, if a chair is detected, \emph{most} chairs have four legs, and this can be inferred from the matched pattern. This matching process can also be used for inferring \emph{behaviors} or \emph{functionalities} of said items, which is even more applicable to this use case.

\section{Conclusion}
\label{sec:conc}
Large language models have quickly become a source of commonsense information, in some cases supplanting search engine use. This easily available source of knowledge can enhance transfer and analogy learning, especially in combination with knowledge engineering methodologies (such as modular ontology modeling).

In this paper, we have described our methodology for obtaining 104 commonsense micropatterns from an LLM. This process is generalizable from our chosen list, and the code to produce additional patterns is available online and permissibly licensed. These patterns are organized into a modular ontology design library (MODL), which we call CS-MODL, also available in the same repository.
CS-MODL can facilitate the construction of internal knowledge frameworks of automated systems through specializing or generalizing from the commonsenese micropatterns.

\subsection*{Future Work}
We have identified some areas for future enhancement.
\begin{compactenum}
    \item The list of nouns that we utilized to seed our prompts for patterns is naïve. That is, no effort was put into organizing them into a subsumption hierarchy or otherwise identifying unifying concepts. Creating both this hierarchy, as well as examining similar hierarchies for relations between these concepts, is an immediate next step. Additionally, these relations can be encoded in the CS-MODL index allowing for substantially more sophisticated pattern identification.
    \item We will evaluate the use of CS-MODL in a more principled way (i.e., through use in a case study) pertaining to ontology learning from text, and in a multi-agent teaming, visual sensor fusion exercise.
    \item Describe the patterns more expressively (i.e., fully describe the patterns in OWL instead of only using RDFS).
    \item Improved documentation: schema diagrams, summaries, and explanation of the formalization (i.e., in the style of \cite{enslaved-jws} and \cite{how2doc}) will be explored.
\end{compactenum}
\medskip

\noindent\emph{Acknowledgement.} The authors acknowledge partial funding by the National Science Foundation under grant 2333532 "Proto-OKN Theme 3: An Education Gateway for the Proto-OKN."
\newpage
\bibliographystyle{splncs04}
\bibliography{refs}
\end{document}

%% file: abstract.tex
\begin{abstract}
The previously introduced Modular Ontology Modeling me\-tho\-do\-logy (MOMo) attempts to mimic the human analogical process by using modular patterns to assemble more complex concepts. To support this, MOMo organizes organizes ontology design patterns into design libraries, which are programmatically queryable, to support accelerated ontology development, for both human and automated processes. However, a major bottleneck to large-scale deployment of MOMo is the (to-date) limited availability of ready-to-use ontology design patterns. At the same time, Large Language Models have quickly become a source of common knowledge and, in some cases, replacing search engines for questions. In this paper, we thus present a collection of 104 ontology design patterns representing often occurring nouns, curated from the common-sense knowledge available in LLMs, organized into a fully-annotated modular ontology design library ready for use with MOMo.
\end{abstract}